\documentclass[letterpaper]{article} 
\usepackage{aaai25}  
\usepackage{times}  
\usepackage{helvet}  
\usepackage{courier}  
\usepackage[hyphens]{url}  
\usepackage{graphicx} 
\usepackage{natbib}  
\usepackage{caption} 
\usepackage{algorithm}
\usepackage{algorithmic}
\usepackage{subcaption}
\usepackage{subfig}
\usepackage{booktabs}
\usepackage{multirow}

\usepackage{newfloat}
\usepackage{listings}
\DeclareCaptionStyle{ruled}{labelfont=normalfont,labelsep=colon,strut=off} 
\floatstyle{ruled}
\newfloat{listing}{tb}{lst}{}
\floatname{listing}{Listing}
\title{\textsc{PokerBench}: Training Large Language Models to \\become Professional Poker Players}
\author{
    Richard Zhuang\textsuperscript{\rm 1}, Akshat Gupta\textsuperscript{\rm 1}\footnote{Correspondence to: akshat.gupta@berkeley.edu}, Richard Yang\textsuperscript{\rm 1}, Aniket Rahane\textsuperscript{\rm 1}, Zhengyu Li\textsuperscript{\rm 2}, \\Gopala Anumanchipalli\textsuperscript{\rm 1}
}
\affiliations{
    \textsuperscript{\rm 1}University of California, Berkeley;
    \textsuperscript{\rm 2}Georgia Institute of Technology\\
}

\usepackage{bibentry}

\begin{document}

\maketitle

\begin{abstract}
We introduce \textsc{PokerBench} - a benchmark for evaluating the poker-playing abilities of large language models (LLMs). As LLMs excel in traditional NLP tasks, their application to complex, strategic games like poker poses a new challenge. Poker, an incomplete information game, demands a multitude of skills such as mathematics, reasoning, planning, strategy, and a deep understanding of game theory and human psychology. This makes Poker the ideal next frontier for large language models. \textsc{PokerBench} consists of a comprehensive compilation of 11,000 most important scenarios, split between pre-flop and post-flop play, developed in collaboration with trained poker players. We evaluate prominent models including GPT-4, ChatGPT 3.5, and various Llama and Gemma series models, finding that all state-of-the-art LLMs underperform in playing optimal poker. However, after fine-tuning, these models show marked improvements. We validate \textsc{PokerBench} by having models with different scores compete with each other, demonstrating that higher scores on \textsc{PokerBench} lead to higher win rates in actual poker games. Through gameplay between our fine-tuned model and GPT-4, we also identify limitations of simple supervised fine-tuning for learning optimal playing strategy, suggesting the need for more advanced methodologies for effectively training language models to excel in games. \textsc{PokerBench} thus presents a unique benchmark for a quick and reliable evaluation of the poker-playing ability of LLMs as well as a comprehensive benchmark to study the progress of LLMs in complex game-playing scenarios.
\end{abstract}

\section{Introduction}
As large language models (LLMs) become exceedingly better at performing traditional natural language processing tasks \citep{glue, superglue, gpt-2, gpt-3}, they are now evaluated on more complicated tasks like recalling world knowledge \cite{MMLU}, reasoning \cite{commonsenseqa} and the ability to mathematics \citep{gsm8k}. A natural next evaluation setting for the ever-growing capabilities of these models is ``game-playing" - a setting that requires not just the ability to do math and reasoning, but also planning, decision-making, and a deeper understanding of opponent behavior and state of mind \citep{games-1, games-2, theory-of-mind}. Poker is one such game that requires the above-described complex skill set.

Poker is an example of an incomplete information game \cite{incomplete-information-games} where a player has complete information about their own holdings, but incomplete information about the holdings of their opponents. The game requires a player to make strategic decisions based on their estimation of the opponent's holdings by accounting for their actions, style of play, game situation, and possible future outcomes \cite{poker-akshat}. This requires a complex combination of skills including math, reasoning, memory, long-term and short-term planning, and strategy, as well as a deep understanding of game theory and player behavior and psychology. Thus, the development of LLMs in game-playing settings like poker can potentially unlock higher cognitive capabilities in these models. 

Existing AI systems for poker, commonly called poker ``solvers", play ``game theory optimal" poker and have been shown to have superhuman performance \citep{poker-noam-counterfactualregret, poker-noam-multiplayer, poker-noam-headsup}. Yet, these solvers have several limitations. (i) Firstly, poker solvers can take a long time to produce solutions for a spot\footnote{A ``spot" in poker is an intermediate point within a game}, thus making them unusable for real-time use. (ii) Secondly, poker solvers are only able to calculate solutions for a limited and discrete set of scenarios, since the game tree in poker\footnote{Here we refer to Texas No-Limit Hold'em version of poker. This version of poker is also the focus of our paper.} can become unmanageably large. The game tree explosion also limits the use of solvers in multi-player settings. (iii) Thirdly, poker solvers do not necessarily provide the most profitable strategies in poker. Solvers are trained to play game theory optimal poker, which means they are trained to be unexploitable. However, being game theory optimal also means that these solvers do not take further advantage when their opponents are playing imperfectly.  (iv) Finally, the solutions or strategies presented by solvers are abstract and not always interpretable which makes it hard to group and study these strategies.

With these limitations in mind and the growing cognitive capabilities of large language models, we explore the use of LLMs as poker solvers. Using LLMs for poker immediately allows for overcoming the disadvantages of traditional poker solvers. LLMs can provide solutions for any spot instantly, and the time taken for a solution is not affected by the number of players or the number of possible scenarios (for example - bet sizes) being considered. LLMs also have the potential to take exploitative actions leading to more profitable strategies by taking into account the user's history of playing using in-context learning \cite{gpt-3}. Finally, since LLMs are text-generation systems, they can also be used to explain their own solutions and strategies. While we do not explore these use cases in this paper, the potential advantages of using LLMs to play poker are exciting. 

As we set out to improve the poker-playing abilities of LLMs, we find a big gap in the form of the absence of a reliable dataset to evaluate the poker-playing ability of LLMs. Poker is a game won or lost in the long term over a large sample of hands. For a strategy in poker to be considered profitable or a player to be considered better, we need to evaluate them over a statistically significant sample of poker hands. This poses a significant challenge in improving LLMs at poker as they need to be evaluated at every step of the way. Running game simulations for tens of thousands of hands after every fine-tuning iteration can significantly slow down this process. Thus, we need a dataset that can be used as a quick and accurate indicator of a model's poker-playing skills. We thus present \textbf{\textsc{PokerBench}} - a new benchmark for a comprehensive evaluation of the poker-playing abilities of LLMs accompanied by a training dataset. The \textsc{PokerBench} benchmark consists of 11,000 spots for evaluating decision-making in poker, covering an exhaustive list of game situations including 1,000 pre-flop scenarios and 10,000 post-flop scenarios. The dataset and code can be found at \url{https://github.com/pokerllm/pokerbench}

In this paper, we make the following contributions:
\begin{itemize}
    \item We introduce the \textsc{PokerBench} benchmark, meticulously developed in collaboration with trained poker players to encompass a full spectrum of essential poker scenarios. Grounded in the rigorous principles of game theory optimal poker, \textsc{PokerBench} evaluates models across a diverse array of important poker spots, encompassing both pre-flop and post-flop play, something that was missing in prior work \cite{poker-akshat}.
    \item Using \textsc{PokerBench}, we perform an extensive evaluation of various state-of-the-art models including GPT-4 \cite{gpt4}, ChatGPT3.5 \cite{chatgpt} the Llama-3 series models \cite{llama3}, the Llama-2 series models \cite{llama2} and Gemma-2B \cite{gemma} on \textsc{PokerBench}. We find that these models seriously underperform in the game of poker compared to their performance on other benchmarks, with GPT-4 being the best-performing model with an accuracy of 53.55\%. We then fine-tune Llama-3-8B, Llama-2-7B, and Gemma-2B models on the accompanying training dataset and find that the performance of the best fine-tuned model improves significantly on fine-tuning. 
    \item To test the usefulness of \textsc{PokerBench} as a quick and reliable measure of the poker-playing ability of LLMs, we have various fine-tuned model checkpoints with different scores on \textsc{PokerBench} play against each other. We clearly see that models with a higher score on \textsc{PokerBench} are able to beat models performing poorly on a large sample of hands.
\end{itemize}

\section{Related Work}
Creating machines that are able to beat humans in gameplay settings has a long history. The first big success for an AI system at beating humans happened in 1997 when IBM's Deep Blue \cite{deepblue} beat Garry Kasparov, the world number 1 chess player and an all-time great of the game. While this was a significant step, Garry Kasparov recovered from a single-game loss and ended up beating Deep Blue 4-2. More recently, AlphaGo \cite{alphago} beat a Go world champion in 2016 in a comprehensive victory of 4-1. Different from Chess and Go, Poker represents a unique challenge by being an incomplete information game \cite{incomplete-information-games}. While in Chess and Go, all pieces of the game are visible to all players; in poker, all information about the opponent's holdings is not available to players. Significant progress was made in creating superhuman AI systems at poker between 2017-2019 with various algorithmic advances \citep{poker-noam-counterfactualregret, poker-noam-headsup, poker-noam-multiplayer}. 

With the increasing cognitive capabilities of LLMs, recent work has started exploring the use of LLMs in gameplay settings. \citet{grandmaster-chess} recently explored the possibility of using LLMs to create non-search-based systems that play ``grandmaster" level chess using LLMs. \citet{poker-akshat} evaluated the poker-playing capability of GPT-4 and ChatGPT, whichrepresents the first work exploring the possibility of playing poker using LLMs. While they concluded that LLMs were not good poker players, their analysis was limited to the first betting action in poker, also called ``raise first in" or RFI spots. \citet{pokergpt} recently fine-tuned LLMs on data collected from PokerStars, a popular poker site, and showed that this improved the poker-playing abilities of LLMs. Yet, most of this development was done in the blind and models could only be evaluated at the end of the fine-tuning process by having them play a large number of hands. \textsc{PokerBench} allows for the constant development of LLMs as poker-playing agents and provides a universal benchmark for creating such systems. 

\section{Poker Preliminaries}
In this paper, we study the most popular version of poker, called Texas No-Limit Hold'em (NLH), which has also been the focus of prior work \citep{poker-akshat, poker-noam-multiplayer, poker-noam-headsup}. The number of players in a Texas NLH game varies from 2-10, with the 6-player game being the most popular. Texas NLH poker is the epitome of decision-making under uncertainty and incomplete information with a near-infinite decision tree. In Texas NLH, each player is given two private cards, also called \textit{hole cards}, that are only known to them, and five community cards that are visible to everyone, peeled in three rounds accompanied by four betting rounds. Each player thus possesses seven cards, out of which two cards are private and only known to the player, and five cards are common to everyone. Out of these seven cards, a player presents their top 5 cards to make a winning combination. The betting rounds are as follows:

\begin{itemize}
    \item \textbf{Pre-Flop}: This is the first betting round that happens right after the players see their hole cards. At this point, none of the five community cards have been opened. 
    \item \textbf{Post-Flop}: After the previous betting round, three community cards are opened at once. This event is called the ``flop". This is followed by a betting round called the post-flop betting round. 
    \item \textbf{Post-Turn}: After the post-flop betting round, a fourth community card is opened. The fourth card is called the ``turn", and is followed by a betting round. 
    \item \textbf{Post-River}: After the post-turn betting round, the fifth and final community card is opened. The final card is called the ``river", followed by a final betting round.
\end{itemize}

A game of poker takes place in multiple iterations which involves distributing hole cards, community cards, and the above-mentioned four betting rounds, resulting in a player winning or losing a pot of chips at stake. One such iteration of the above process is commonly referred to as a  ``\textit{hand}".

\subsection{Actions in Poker}
 There are four basic actions that a player takes while playing poker :

\begin{itemize}
    \item \textbf{Check}: This action means that a player wants to continue playing in a betting round without wagering any chips. This can only be done if no wagers have yet been made previously in that betting round. 
    \item \textbf{Bet}: The action of placing a wager in a betting round is called a \textit{bet}. In poker, ``\textit{bet}" is specifically referred to a situation when no wager has yet been made in a betting round, then the first wager is called a ``\textit{bet}".
    \item \textbf{Call}: The action of matching a wager made by a player previously is called a ``\textit{call}". 
    \item \textbf{Raise}: The action of wagering a larger amount of chips than the previous wager is called ``\textit{raise}".
    \item \textbf{Fold}: The action of choosing not to match a previous wager (bet or raise) and thus giving up claim on the pot is called a ``\textit{fold}". When a player folds, they are no longer part of the current \textit{hand} being played. 
\end{itemize}

\subsection{The Unit of Measurement in Poker}
Poker is a very popular game played at different stakes. While the lowest stakes in most casinos require a buy-in of 100\$, the buy-in can go as high as millions of dollars in high-stakes games around the world. Yet, every person in the game usually starts with the same amount of effective stack size\footnote{``stack size" is the total number of chips a player has} according to a normalized unit of measurement. 

The normalized unit of measurement in poker is the minimum amount of money a player is allowed to bet, and is called a ``\textbf{big blind}" (BB). If the minimum bet amount in a game is 2\$, then 1BB = 2\$. If the minimum bet amount in a game is 200\$, then 1BB = 200\$. Poker games are usually classified by the amount required to buy 100 big blinds. So if a game is classified as a 100\$ buy-in game, this means that the minimum bet is 1\$. A standard starting stack size for most poker games and solver calculations is 100BB. 

Throughout this paper, we will be describing gameplay using the BB unit. The win rate of a poker player is also defined in terms of big blinds, as the number of big blinds won per hundred hands played.

\subsection{Game Theory Optimal Poker}\label{subsec: GTO}
``Game Theory Optimal" (GTO) strategy of playing poker refers to the optimal way of playing poker such that a player cannot be exploited by their opponents. A GTO strategy is usually a balanced strategy where the opponent is unable to correlate the actions of a player with their holdings. For a simplistic pedagogical example to illustrate this, if Player A's strategy is to go all-in with only pocket aces (\texttt{AA}), which are the best starting hands in poker, then Player B can exploit Player A by folding against Player A's all-in with any two cards that are not pocket aces (\texttt{AA}). As a result, Player A is unable to extract value from their strongest hand. Thus, in GTO play, Player A should go all-in with a wider selection of hands, and also not go all-in with pocket Aces every single time. This is a more optimal strategy in the long run against all kinds of players since it becomes challenging for our opponent to narrow down player A's exact holding. A widely used technical term for the philosophy behind such a strategy is called playing a ``balanced'' game.

\section{The \textsc{PokerBench} Benchmark}
We carefully design the \textsc{PokerBench} benchmark to thoroughly evaluate the poker-playing abilities of LLMs with an exhaustive coverage of many types of poker spots. The aim of creating \textsc{PokerBench} is to evaluate LLMs at playing game theory optimal poker in a quick and reliable way. \textsc{PokerBench} is designed such that the higher a model scores on our benchmark, the better it is at playing optimal poker. 

We define a ``spot" to be a combination of hole cards that a player has, the board\footnote{``\textit{the board}" is a term used to refer to the set of community cards in play in a particular hand since they traditionally get displayed on a board}, and the actions taken in the different betting rounds. Similar to chess, the search space for Texas NLH poker is extremely large. Thus, we build \textsc{PokerBench} with two main balancing principles: diversity and simplicity. We want to be able to evaluate an LLM thoroughly by having them play a wide category of scenarios while keeping a reasonable total inference time to enable quick development. The following subsections explain our design choices in condensing the enormous search space by careful filtering and pruning using principles of optimal poker play.

The \textsc{PokerBench} benchmark is created based on 6-max player Texas NLH poker. It consists of two separate sets - a pre-flop and a post-flop evaluation dataset. Pre-flop games are usually very different from post-flop play and have a very different distribution of decisions, which is why we decided to separate the two types of scenarios. We have 1,000 evaluation examples in the benchmark for the pre-flop game and 10,000 evaluation examples for the post-flop game. To create this dataset, we use the GTO strategies from GTOWizard\footnote{\url{https://github.com/mtpham99/gtowizard_scrape_public}} for the pre-flop game and WASM-Postflop\footnote{\url{https://github.com/b-inary/wasm-postflop}} to solve GTO strategies for the post-flop game.

\subsection{Pre-flop Action Selection}

After each player is dealt with two hole cards, the first betting round can have an exponentially large number of decisions being made. For example, a player can decide to bet a certain amount, and a following player can decide to bet an even larger amount (called a ``raise" in poker), and this raising and re-raising can happen with an exploding number of permutations, each with a different bet and raise size. Existing pre-flop GTO strategies exist for all of these scenarios. In \textsc{PokerBench}, we only consider scenarios where a maximum of two raises have happened in the pre-flop betting round. This includes scenarios where (i) all players fold, (ii) only one player bets chips and other players either call or fold, and (ii) one player bets chips, a second player raises that bet by a higher wager followed by only calls or folds (also called ``3-bet pots"). This covers the majority of the possible pre-flop scenarios that are considered a viable GTO play as most pots do not go beyond a single raise.

\subsection{Board Selection}
A ``board" is the list of community cards that show up during a poker game. There are in total 5 community cards, where each card can be any of the 52 cards in the deck. Thus, we have a total of $^{52}C_5 \approx 311$ million possible boards that can show up. It is impossible to evaluate our model on that many boards, which is why we group the boards into 11 classes, called \textbf{textures}. These board textures cover the most common situations on the flop.

One of the most commonly studied board textures on the flop is what we call ``single-broadway-dry". As a reminder, ``flop" is a term used to describe the first three community cards. The term ``broadway cards" is used to refer to the cards in the set \texttt{\{A,K,Q,J,10\}}, that is, it refers to the five strongest cards in a suit. The term  ``single-broadway" refers to the opening of exactly one broadway card on the flop. The term ``dry" is used to describe flops in poker which do not have a lot of possibilities to make different winning hands. For example, a board \texttt{\{K$_h$,7$_d$,2$_s$\}}, where the subscripts show the suit of the cards\footnote{h : hearts, d: diamonds, s: spades, c: clubs}, is a typical dry board in poker. The board does not have any repetition of suit, which means it is less likely to form flushes\footnote{when all five cards have the same suit}, and does not have numbers close to each other, which means it is less likely to form straights\footnote{a consecutive sequence of 5 cards}. The strategy a player should use to play on a \texttt{\{K$_h$,7$_d$,2$_s$\}} board is going to be very similar to strategy on similar boards like \texttt{\{K$_h$,7$_d$,4$_s$\}}, \texttt{\{A$_h$,8$_d$,3$_s$\}}, \texttt{\{Q$_h$,8$_d$,2$_s$\}}, and hence all such boards are group together into a single texture called ``single-broadway-dry". All board textures used to create \textsc{PokerBench} are shown in appendix \ref{sec:textures}. 

We randomly sample an equal number of flops from each of the textures to create the dataset of boards on which the game is evaluated. Then to select the turn cards for each scenario, we select cards that would continue to cover the most new ground. For example, if a flop contains a flush draw\footnote{A ``flush draw" exists when two cards of the same suit are opened on the board, thus making the possibility of a flush if another card of the same suit is dealt}, some possible turn cards were selected to either complete or not complete the flush draw. This process ensures an equal coverage of the different categories of boards and results in a more informed evaluation compared to random sampling, especially at small sample sizes.

\begin{table}
\scalebox{0.88}{
    \centering
    \begin{tabular}{p{2.5cm}|p{2.3cm}|p{2.3cm}} 

    \textbf{DATASET TYPE} & \textbf{PRE-FLOP SPOTS} & \textbf{POST-FLOP SPOTS} \\ \hline\hline
    \textsc{PokerBench} & 1,000 & 10,000 \\ \hline
    Training Set & 60,000 & 500,000\\
    \end{tabular}
}

    \caption{\textsc{PokerBench} summary}
    \label{tab:dataset-summary}
\end{table}

\subsection{Selecting Hole Cards}
GTO play dictates to have nondeterministic strategies since if a deterministic strategy exists in a player's game, it becomes exploitable. For example, if a player always goes all-in when they have aces (\texttt{AA}), then all other players are likely to fold, ending up in an unprofitable play. Thus, a balanced, game theory optimal way of playing any spot in poker is to have at least two actions chosen with certain probabilities at the time of play. For selecting hole cards, we choose spots where there is a clear dominant strategy as the best action. For example with a board like \texttt{\{A$_h$,K$_h$,8$_d$,3$_s$,10$_c$\}}, facing a small bet from the opponent, private cards that have a clear dominant strategy are hands like \texttt{Q$_d$,J$_s$} of raising, whereas hands like \texttt{10$_d$,9$_d$} have a more mixed strategy of calling or folding. We decided to filter hole cards by selecting action lines that choose one dominant action with greater than 50\% probability.

        
        
        

\begin{figure*}[t]
    \centering
    \begin{minipage}[t]{0.98\textwidth}
        \centering
        \begin{subfigure}[t]{0.24\textwidth} 
            \centering
            \includegraphics[width=\textwidth]{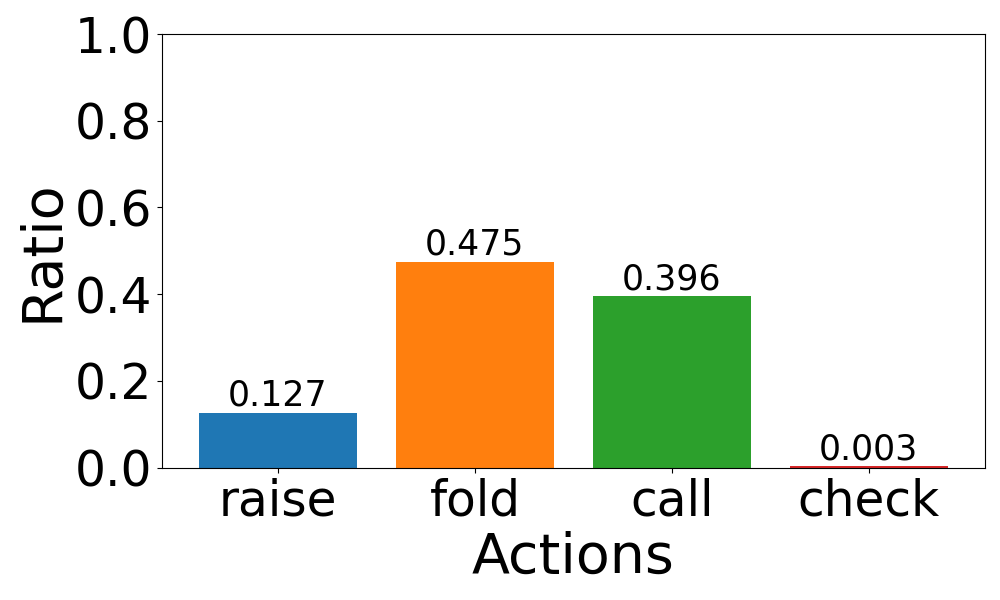}
            \caption{Pre-flop Train Set}
        \end{subfigure}%
        \hfill
        \begin{subfigure}[t]{0.24\textwidth} 
            \centering
            \includegraphics[width=\textwidth]{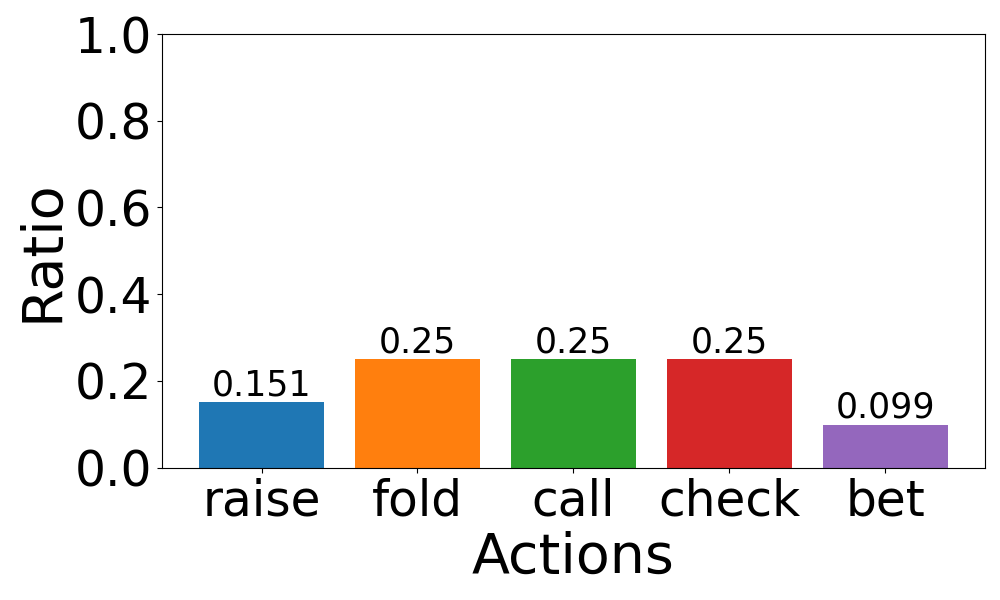}
            \caption{Post-flop Train Set}
        \end{subfigure}%
        \hfill
        \begin{subfigure}[t]{0.24\textwidth} 
            \centering
            \includegraphics[width=\textwidth]{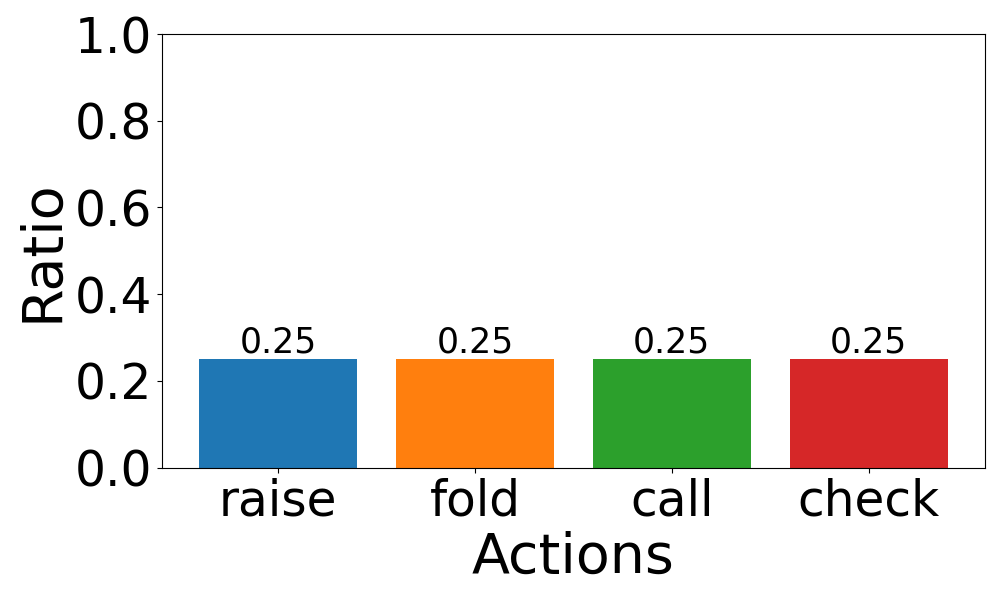}
            \caption{Pre-flop Benchmark}
        \end{subfigure}%
        \hfill
        \begin{subfigure}[t]{0.24\textwidth} 
            \centering
            \includegraphics[width=\textwidth]{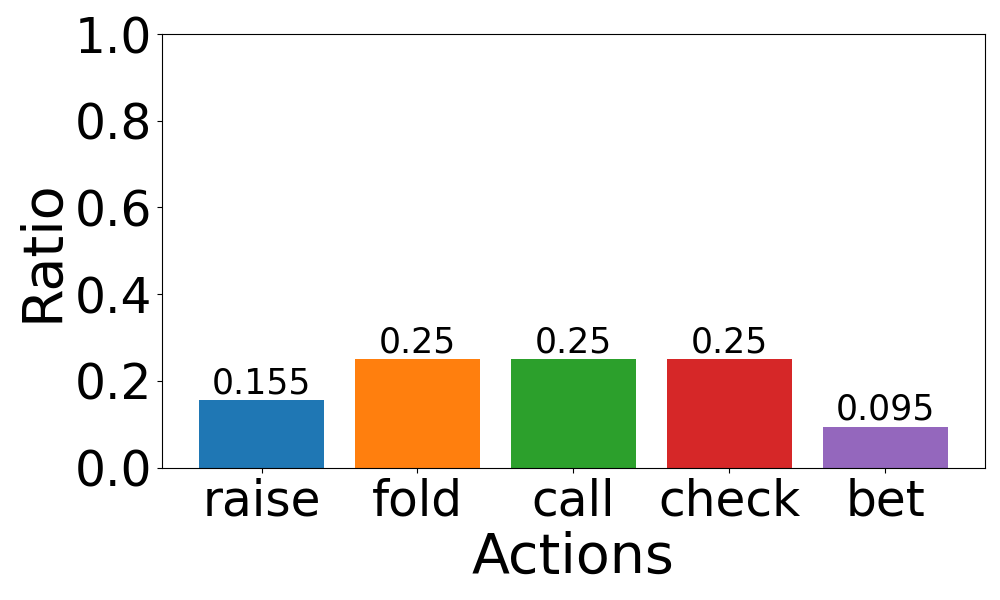}
            \caption{Post-flop Benchmark}
        \end{subfigure}
        
        \caption{Action distributions. (a) and (b) show the training dataset distribution, while (c) and (d) show the distribution for hands selected to be part of the \textsc{PokerBench} Benchmark.}
        \label{fig:dataset-distributions}
    \end{minipage}
\end{figure*}

\subsection{Dataset Summary}
The \textsc{PokerBench} benchmark consists of 1k evaluation spots for pre-flop and 10k evaluation spots for post-flop play as shown in Table \ref{tab:dataset-summary}. Along with the evaluation benchmark, we also release a training set containing 60k pre-flop spots and 500k post-flop spots. The distribution of actions for the \textsc{PokerBench} benchmark and the training set can be seen in Figure \ref{fig:dataset-distributions}.

For the pre-flop training set, we preserve the original distribution of actions that a player can take because the search space there is relatively small, and so an exhaustive strategy that covers most possible scenarios is learnable. For the post-flop training set, we experiment with a few different sampling strategies and find that resampling the action distribution using a balanced strategy results in the best fine-tuning performance. Specifically, a balanced sample helps ensure coverage of the search space. It also keeps the model from learning a naive strategy that decides to ``fold" most of the time, a possible situation if we preserve the original distribution where the action of folding predominates. 

However, for the \textsc{PokerBench} benchmark, for both pre-flop and post-flop settings, we choose a balanced sampling strategy. Precisely, there is an equal percentage of samples each with the correct decision labels being fold, call, check, or bet/raise. We cannot preserve the highly imbalanced distribution for evaluation because a naive strategy of folding all hands will lead to an accuracy of close to 90\%. Because of this, we downsample the fold action significantly to remove hands that do not provide much signal. For example, all starting hands like \texttt{\{72\},\{83\},\{93\},\{84\}} should be folded, but not all of these hands need to be kept in the evaluation set to get the signal that the model is folding such low-value hands. Whereas hands like \texttt{\{AK\},\{KQ\},\{AA\},\{JJ\}} should be played before the flop, it is important to analyze the model on a larger set of such hands, as these hands are also more profitable.

\begin{table*}
    \vskip 0.05in
    \centering 
    \scriptsize
    \setlength\tabcolsep{0pt}
    \begin{tabular*}{\textwidth}{@{\extracolsep{\fill\centering}}*{8}{c}}
        \toprule 
        \multirow{2}{*}[-0.3em]{\textsc{Evaluation Type}} & 
            \multirow{2}{*}[-0.3em]{\textsc{Model}} &
            \multicolumn{2}{c}{Overall Accuracy} & 
            \multicolumn{2}{c}{Post-Flop Accuracy} & 
            \multicolumn{2}{c}{Pre-Flop Accuracy}\\ 
        \addlinespace[0.125em] \cline{3-8} \addlinespace[0.25em]
        &  & EM $\uparrow$ & AA $\uparrow$ &
             EM $\uparrow$ & AA $\uparrow$ &
             EM $\uparrow$ & AA $\uparrow$\\
        \midrule 
        \multirow{2}{2cm}[-0em]{\centering Pre-Trained Models (Few-Shot)} & \textsc{Llama-3 (8B)} & $26.02$ & $40.03$ & $14.96$ & $31.25$ & $37.77$ & $49.30$\\
        & \textsc{Llama-2 (70B)}& $36.48$ & $48.30$ & $32.95$ & $41.11$ & $40.20$ & $55.90$\\
        & \textsc{Llama-3 (70B)}& $39.16$ & $49.78$ & $34.30$ & $45.40$ & $44.30$ & $54.40$\\
        & \textsc{ChatGPT 3.5}& $29.96$ & $39.69$ & $18.75$ & $34.19$ & $41.80$ & $45.50$\\
        & \textbf{\textsc{GPT-4}}& $\mathbf{53.55}$ & $\mathbf{65.54}$ & $\mathbf{52.18}$ & $\mathbf{62.69}$ & $\mathbf{55.00}$ & $\mathbf{66.50}$\\
        \midrule
        
        \multirow{2}{2cm}[-0em]{\centering Fine-Tuned Models (Zero-Shot)} & \textsc{Gemma (2B)} & $51.84$ & $62.74$ & $41.57$ & $52.94$ & $62.70$ & $73.10$ \\
        & \textsc{Llama-2 (7B))}&  $78.11$ & $79.91$ & $76.52$ & $79.55$ & $79.80$ & $80.30$\\
        & \textbf{\textsc{Llama-3 (8B)}}& $\mathbf{78.26}$ & $\mathbf{80.64}$ & $\mathbf{76.52}$ & $\mathbf{79.07}$ & $\mathbf{80.10}$ & $\mathbf{82.30}$\\
        \midrule\bottomrule
    \end{tabular*}
        \caption{Performance of various pre-trained and fine-tuned LLMs on \textsc{PokerBench}. }\label{table:main-results}
    \vskip -0.0in
\end{table*}

\section{Experiments}
After carefully curating \textsc{PokerBench}, we move on to evaluating the poker-playing ability of LLMs. 

\paragraph{Evaluation Metrics}
We use the following two metrics to evaluate the poker skills of LLMs by evaluating their responses on \textsc{PokerBench}: 

\begin{itemize}
    \item \textbf{Action Accuracy (AA):} Action accuracy measures if LLMs can take game theory optimal actions (fold, raise, bet, etc.) for a given spot. 
    \item \textbf{Exact Match Accuracy (EM):} Actions like \textit{bet} and \textit{raise} are followed by the amount of the bet and raise. Hence, exact match accuracy also considers if the wager amount is game theory optimal. 
\end{itemize}

\subsection{Are Modern LLMs Good at Playing Optimal Poker?}\label{sec:pretrained}
We evaluate GPT-4 \cite{gpt4}, ChatGPT 3.5 \cite{gpt3.5}, Llama-3 models (8B, 70B) \cite{llama3} and Llama-2 70B \cite{llama2} on \textsc{PokerBench}. We take inspiration from the evaluation protocol for the popular MMLU dataset \cite{MMLU} and evaluate these models in a few-shot setting. While MMLU is a multiple-choice question-answering dataset, \textsc{PokerBench} is not. Thus we require the model to generate the action and the exact bet amount for a given spot. For few-shot examples, we select one example randomly for each possible action from the training dataset and add it to the context. Thus our few-shot setting contains 5 examples in context, one for each action as shown in Figure \ref{fig:dataset-distributions}. An example prompt can be found in Table \ref{table:prompt} (appendix). For generating text, we set temperature = 0.1 and top-p = 0.95 to generate the most probable answer to get statistically stable results. We use the OpenAI API\footnote{\url{https://platform.openai.com/}} for evaluating OpenAI models and TogetherAI API\footnote{\url{https://docs.together.ai/docs/quickstart}} for evaluating models from the Llama series.

The results for evaluation on \textsc{PokerBench} can be found in Table \ref{table:main-results}. We use chat or instruct models for this evaluation as applicable\footnote{Available open-sourced base models performed sub-optimally compared to chat/instruct models which is why we do not report the scores of the base models.}.  GPT-4 outperforms all other models both in pre-flop and post-flop play. The second best performing model is Llama-3-70B with its performance on the benchmark being significantly lower than GPT-4. A surprising thing to note is that ChatGPT 3.5 performs comparably to a significantly smaller Llama-3-8B model. Llama-3 outperforms Llama-2 in generating correct bet/raise amounts (higher EM score) while this lead diminishes when generating optimal actions (similar AA score).  While we also tried to evaluate the smaller models from the Llama-2 series, they are unable to follow poker instructions, which is why we do not report their accuracy scores. The benchmark results show that all modern LLMs significantly lack in their ability to play game theory optimal poker and that there is a lot of room for improvement.

\subsection{Fine-Tuning LLMs into Better Poker Players}\label{sec:sft}
In the previous section, we saw that state-of-the-art LLMs are not good at playing poker and significantly underperform on \textsc{PokerBench} compared to other tasks they are evaluated on \citep{MMLU, commonsenseqa, glue, superglue}. A possible reason for this could be the complexity of the game which requires a multitude of skills coming together. To improve the poker-playing ability of LLMs, we fine-tune the model on a subset of the accompanying training set released with this paper. The training set consists of 30k pre-flop spots and 80k post-flop spots. We have a much larger amount of post-flop spots in the training set since there are more possible permutations of post-flop spots compared to the pre-flop game. We choose this subset to balance the number of examples in the pre-flop and post-flop stages. 

\begin{figure}
    \centering
    
    \centering
    \includegraphics[width=0.48\textwidth]{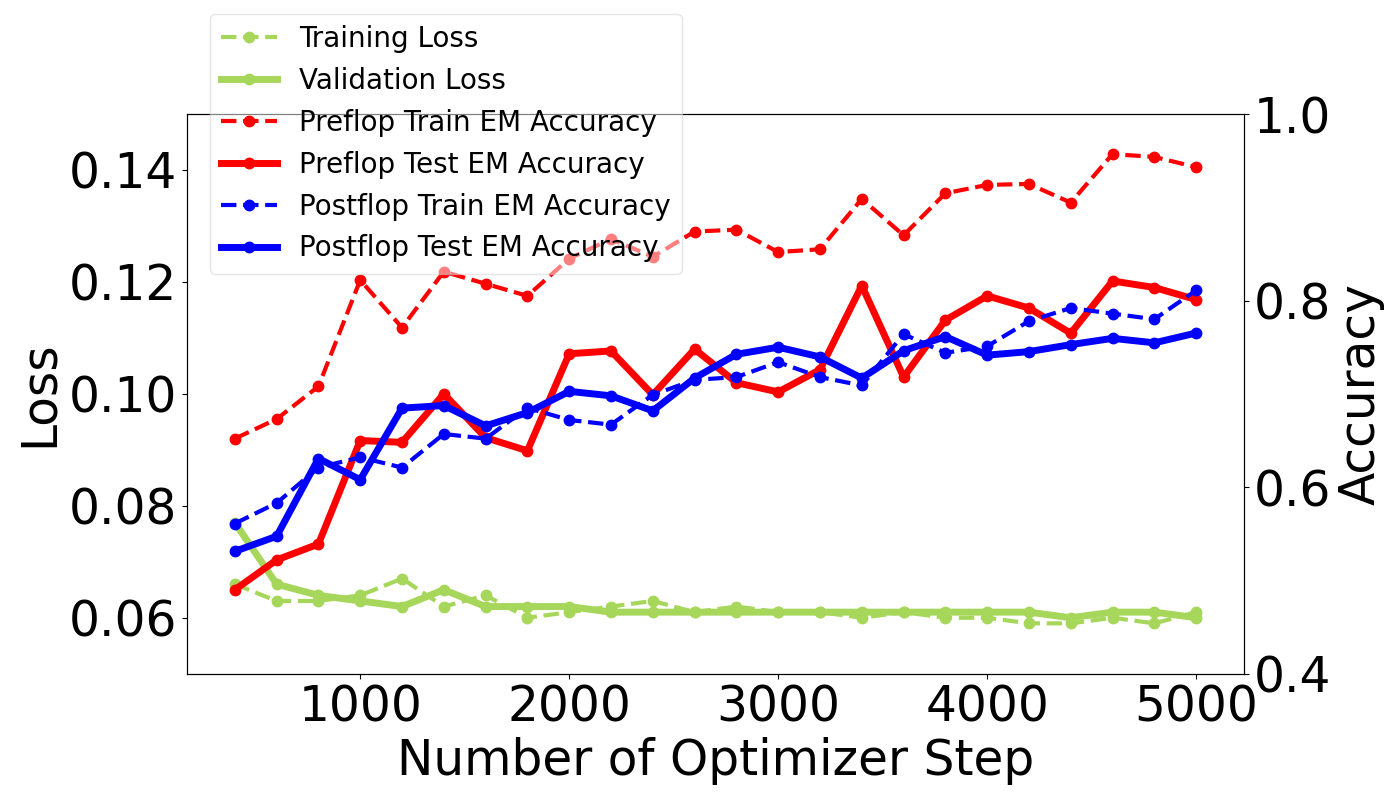}
    \caption{Training dyanmics of Llama-3-8B on \textsc{PokerBench} training dataset.}\label{fig:sft-llama-3}
\end{figure}

We fine-tune three models on the \textsc{PokerBench} training dataset - Llama-3-8B, Llama-2-7B, and Gemma-2B. We fine-tune the model for 5000 optimization steps with a batch size of 128 and a learning rate of 1e-6, which is lower than the pre-training learning rate for most of these models\footnote{A higher learning rate resulted in unstable fine-tuning}. The model goes through one epoch of the dataset approximately every 900 gradient steps. The performance of the fine-tuned models on \textsc{PokerBench} can be seen in Table \ref{table:main-results}. We see that the Llama-3-8B model improves considerably on the \textsc{PokerBench} benchmark, outperforming GPT-4. The performance for Llama-3-8B and Llama-2-7b are quite similar with Llama-3-8B performing slightly better, whereas the Gemma-2B model falls behind on the benchmarks. Thus going forward, we pick Llama-3-8B for further investigation. Figure \ref{fig:sft-llama-3} shows the loss curves for Llama-3-8B plotted with the accuracy of different checkpoints during the fine-tuning process. The loss curves for Llama-2-7B and Gemma-2B are presented in Figures \ref{fig:sft-llama-2} and \ref{fig:sft-gemma} in the appendix and have very similar training dynamics. 

\paragraph{Heads-Up Games between Fine-tuned Checkpoints}

Finally, we test if the scores on \textsc{PokerBench} translate into actual performance. To do so, have different models with different scores play an actual poker game against each other. To do so, we pick three different Llama-3-8B model checkpoints with different accuracy scores on \textsc{PokerBench}. We simulate a heads-up\footnote{A ``heads-up" match in poker is a game where two players play against each other.} game between the three players. The details of the selected players are shown in Table \ref{tab:checkpoints}. The checkpoint name refers to the number of gradient updates that have happened in the fine-tuning process at the time of selection. For example, the Llama-3-8B 1600 has gone through 1600 gradient update steps on the training dataset. Going forward, we will refer to the players by their checkpoint names as player 800, player 1600, and player 5000. 

\begin{table}
\scalebox{0.90}{
    \centering
    \begin{tabular}{p{4cm}|p{1cm}|p{1cm}} 

    \textbf{CHECKPOINT NAME} & \textbf{EM} & \textbf{AA} \\ \hline\hline
    \centering Llama-3-8B 800 & 58.62 & 61.32 \\ \hline
    \centering Llama-3-8B 1600 & 65.96 & 69.60 \\ \hline
    \centering Llama-3-8B 5000 & 78.26 & 80.64 \\ \hline
    \end{tabular}
}

    \caption{Selected checkpoints for heads-up games. }
    \label{tab:checkpoints}
\end{table}

\begin{table}
\scalebox{0.90}{
    \centering
    \begin{tabular}{p{2.1cm}|p{1.5cm}|p{3.5cm}} 

    \centering \textbf{GAME} & \textbf{WINNER} & \textbf{WIN RATE} \\ \hline\hline
    \centering 5000 vs 1600 & 5000 & 24.79 bb/100\\ \hline
    \centering 5000 vs 800 & 5000 & 50.88 bb/100 \\ \hline
    \centering 1600 vs 800 & 1600 & 34.99 bb/100 \\ \hline
    \end{tabular}
}

    \caption{Results of heads-up games against players 5000, 1600 and 800. }
    \label{tab:tournament-results}
\end{table}

The heads-up games are played with two modifications - the stacks of the players are always reset to 100BB to match the GTO strategies\footnote{GTO strategies are calculated at 100BB stacks}, and the seating positions of the players are randomly assigned before each hand. To create statistically significant results, we have each player play against the other for 50k hands. The win rate is calculated with a metric called big blinds per hundred hands, or bb/100, defined as the number of big blinds won per hundred hands played. As a reminder, a big blind is the basic unit of measurement in poker and represents the minimum bet that can be made. As a reference, under a statistically significant sample size, a win rate of about 4.0 bb/100 is considered very good in poker. A win rate in the range of 5.0-9.0 bb/100 is considered exceptional and is a sign of a clearly dominant player. At this win rate, players are suggested to move at a higher level of stake.

The results for the heads-up tournament can be seen in Table \ref{tab:tournament-results}. We see that player 5000, who also has the highest score on \textsc{PokerBench}, is a significantly superior player compared to the other players. Player 5000 wins its games against players 1600 and 800, while the same is true for player 1600 in its game against player 800. These experiments show that a higher score on \textsc{PokerBench} actually translates to a higher win rate. These experiments also present the value of \textsc{PokerBench} as an evaluation tool during development. Instead of having models play against each other for 50k hands for statistical significance, we can evaluate the models on a much smaller set of examples chosen from \textsc{PokerBench}.

\begin{table}
\scalebox{0.90}{
    \centering
    \begin{tabular}{p{2.5cm}|p{1.5cm}|p{3.5cm}} 

    \centering \textbf{GAME STAGE} & \textbf{WINNER} & \textbf{WIN RATE} \\ \hline\hline
    \centering Overall & GPT-4 & 22.20 bb/100\\ \hline
    \centering Pre-flop & 5000 & 17.80 bb/100\\ \hline
    \centering Flop & GPT-4 & 34.97 bb/100 \\ \hline
    \centering Turn & GPT-4 & 24.04 bb/100 \\ \hline
    \centering River & 5000 & 18.02 bb/100 \\ \hline
    \end{tabular}
}

    \caption{Results of the 1,000 heads-up games between Player 5000 and GPT-4. As the sample size is limited, the absolute value of win-rate is less meaningful. Instead, we provide a playing style analysis to better understand the models' performances.}
    \label{tab:tournament-results-gpt4}
\end{table}

\paragraph{Fine-tuned Model vs. GPT-4}

Next, we perform a heads-up game between the best fine-tuned model checkpoint and the best pre-trained model, namely GPT-4, using a same setting. Due to inference cost constraints, the two models played for 1,000 hands only. The result for this tournament can be seen in Table \ref{tab:tournament-results-gpt4}. 

We notice an unorthodox phenomena, that despite achieving higher test accuracy, our fine-tuned Llama model was outplayed by GPT-4. To investigate this, we separate the win-rate by stage of the game (pre-flop, flop, turn, and river), and conduct an in-depth analysis of the game log. Overall, we identify spots where Llama is effectively learning from the training set and winning by playing more GTO than GPT-4. For instance, as shown in Table \ref{tab:gameplay_freq_analysis}, Llama is winning in the pre-flop stage by raising a wider range of hands (which is more optimal) and squeezing GPT-4 out of the pot. 

However, one reason of GPT-4's winning is that it adopts a sub-optimal strategy called ``donking". In a nutshell, it is a losing move in the long run, hence not frequently seen in GTO strategies (more details can be found in the Appendix). But our dataset assumes GTO play from both players. Consequently, this move is unfamiliar to the fine-tuned model, constituting a big part of Llama's losses as GPT-4 wins 100\% of the hands where it donks. 


Also, neither player is playing at an optimal action frequency: GPT-4 is being over-aggressive while Llama is over-passive. A good example of this can be illustrated by a rock-paper-scissors game. Imagine player A chooses rock 100\% of time, while player B chooses scissors 50\% of time and rock 50\% of time. We can see that even though player B has a strategy closer to the optimal solution\footnote{The optimal strategy in rock-paper-scissors is playing each option 33\% times.}, it will never win against player A's less optimal strategy. Similarly, while GPT-4 has a strategy farther from GTO (as assessed by our dataset), it is still able to win against Llama by coincidentally executing a strategy that happens to exploit Llama's suboptimal decisions. A more detailed analysis can be found in the Appendix. 

\paragraph{Discussion} 
 There is a fine line between winning against an imperfect strategy versus minimizing losses against a GTO strategy. The two methods of evaluation used in this paper are evaluating two different things. The gameplay is evaluating how well a strategy performs against an imperfect strategy, while our dataset is evaluating the how well a strategy perform against a GTO strategy. One of the advantage about training the model to approach GTO strategy is that being GTO is a sufficient condition for winning against any imperfect strategy. But when an non-perfect GTO strategy is learned (for instance in the case of our fine-tuned model since it is not perfectly GTO), the relationship between the two could become complex. Yet, we argue that succeeding on our dataset is a necessary condition for minimizing loss against a GTO strategy because our dataset is a compilation of a diverse set of samples that any near GTO strategy must agree on (and non-GTO strategy is guaranteed to lose in the long run). We also believe that this finding is also a limitation of the simple supervised fine-tuning in training LLMs to effectively succeed in poker. We would like to use this example to encourage further research in improving the adaptability of language models in game environments.
 

\section{Conclusion}
In this paper, we present \textsc{PokerBench} - a comprehensive benchmark that evaluates optimal poker-playing ability of LLMs. We evaluate multiple state-of-the-art language models and show that current LLMs fail significantly in playing optimal poker. We also fine-tune Llama-3-8B among other models on the accompanying training dataset and show that resultant models can outperform much larger models. We also show that the scores on \textsc{PokerBench} actually translate to superior poker-playing skills by evaluating models of different scores through game simulations over 50k hands. Thus, \textsc{PokerBench} represents a quick and reliable measure of the optimal poker-playing ability of large language models as well as a comprehensive benchmark to study the progress of LLMs in this domain. This study not only explores the potential of LLMs in strategic game-playing but also presents a benchmarks to evaluate higher-level cognitive capabilities of LLMs in complex game-playing situations.

\newpage
\bibliography{aaai25}

\appendix

\newpage

\section{Appendix}
\subsection{Explanation of Board Textures}\label{sec:textures}
\begin{figure*}
    \centering
    
    \begin{subfigure}[b]{0.2\textwidth}
        \centering
        \includegraphics[width=\textwidth]{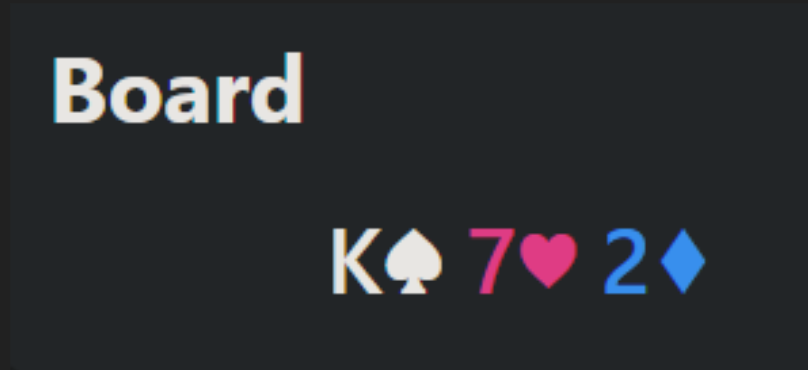}
        \caption{Single Broadway Dry}
    \end{subfigure}
    \hfill
    \begin{subfigure}[b]{0.215\textwidth}
        \centering
        \includegraphics[width=\textwidth]{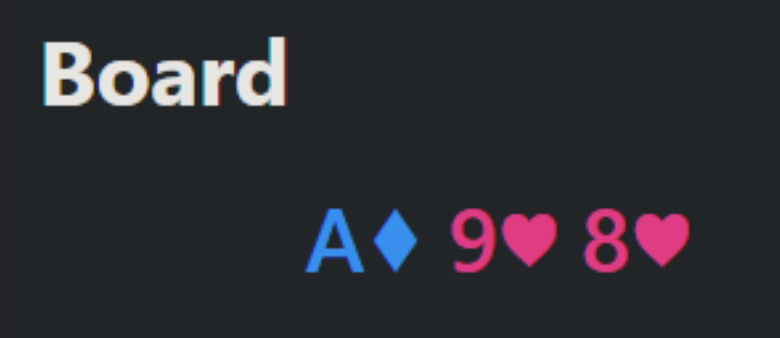}
        \caption{Single Broadway Wet}
    \end{subfigure}
    \hfill
    \begin{subfigure}[b]{0.2\textwidth}
        \centering
        \includegraphics[width=\textwidth]{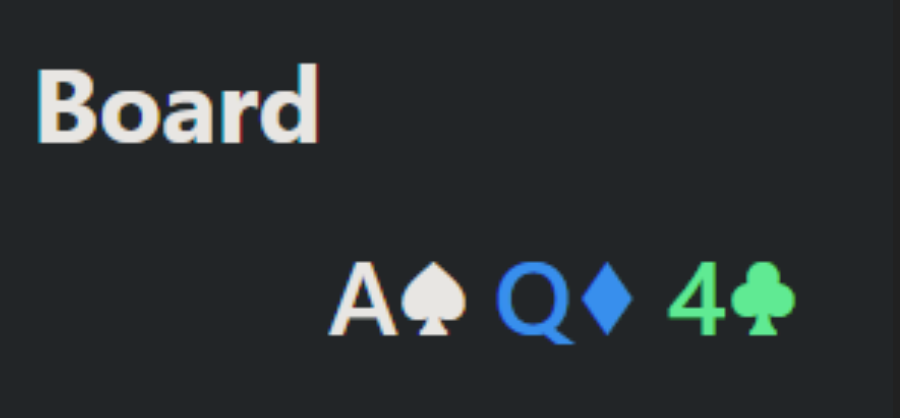}
        \caption{Double Broadway}
    \end{subfigure}
    \hfill
    \begin{subfigure}[b]{0.2\textwidth}
        \centering
        \includegraphics[width=\textwidth]{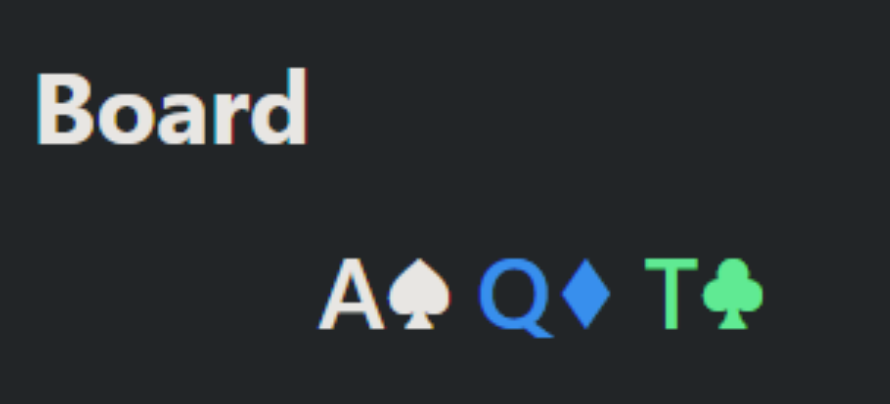}
        \caption{Triple Broadway}
    \end{subfigure}\\
    \vfill
    \begin{subfigure}[b]{0.2\textwidth}
        \centering
        \includegraphics[width=\textwidth]{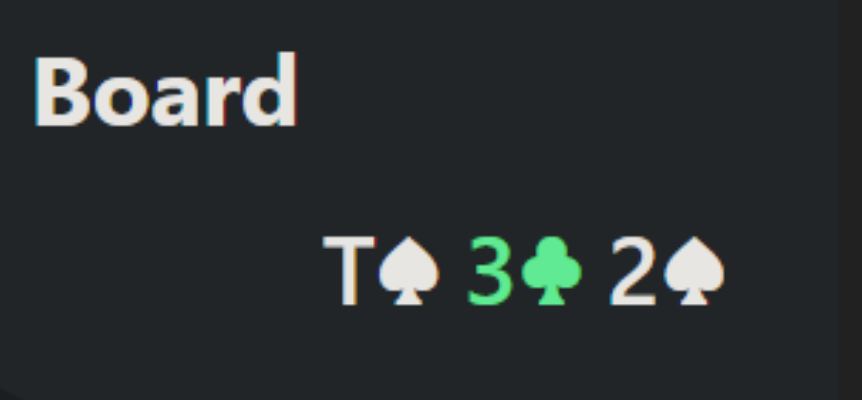}
        \caption{Mid Dry}
    \end{subfigure}
    \hfill
    \begin{subfigure}[b]{0.2\textwidth}
        \centering
        \includegraphics[width=1\textwidth]{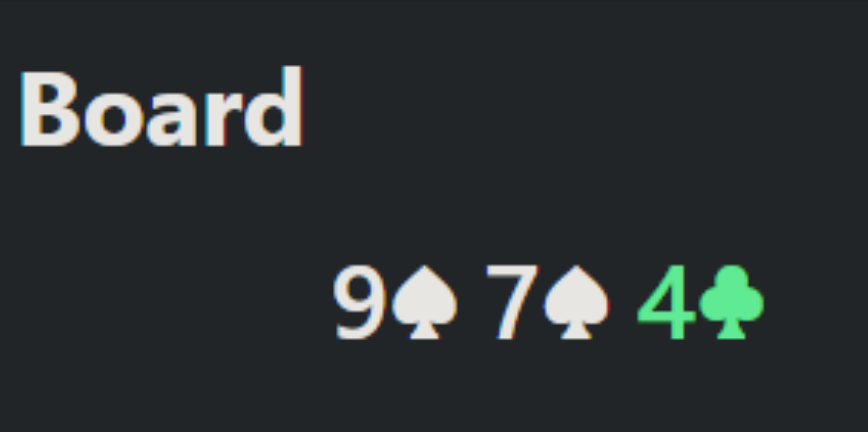}
        \caption{Dynamic}
    \end{subfigure}
    \hfill
    \begin{subfigure}[b]{0.2\textwidth}
        \centering
        \includegraphics[width=\textwidth]{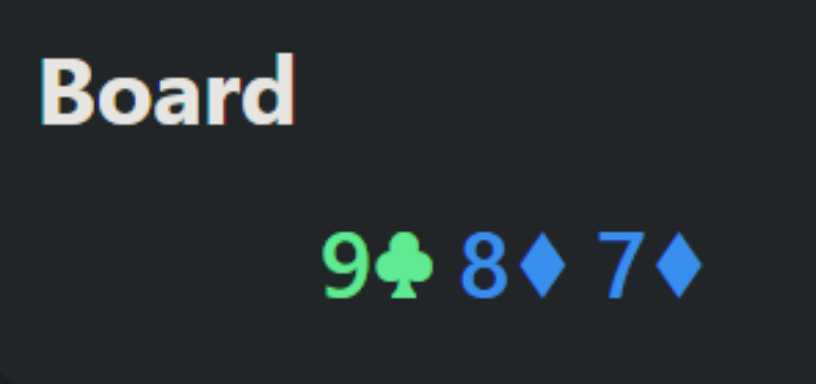}
        \caption{Very Wet/Dynamic}
    \end{subfigure}
    \hfill
    \begin{subfigure}[b]{0.2\textwidth}
        \centering
        \includegraphics[width=\textwidth]{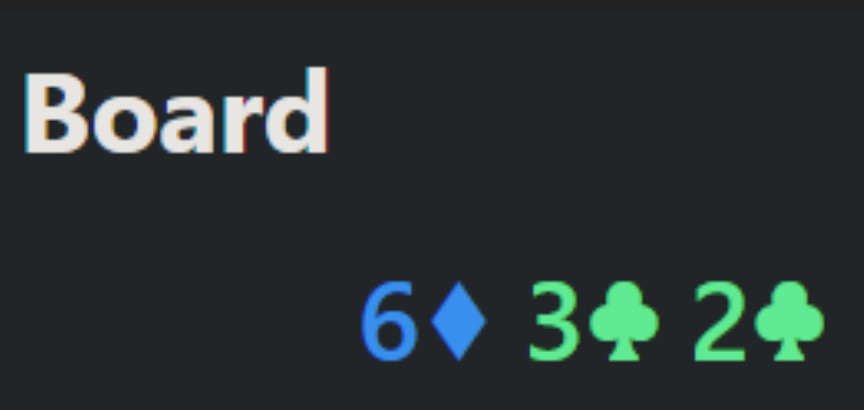}
        \caption{Low}
    \end{subfigure}\\
    \vfill

    \begin{subfigure}[b]{0.2\textwidth}
        \centering
        \includegraphics[width=\textwidth]{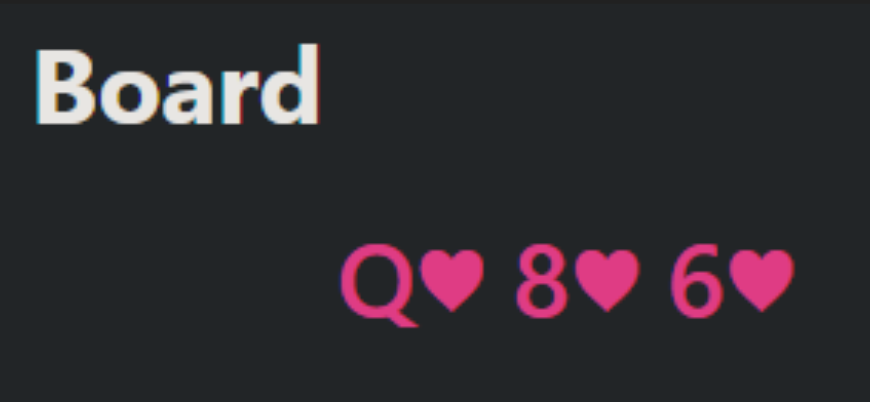}
        \caption{Monotone}
    \end{subfigure}
    \hfill
    \begin{subfigure}[b]{0.2\textwidth}
        \centering
        \includegraphics[width=1\textwidth]{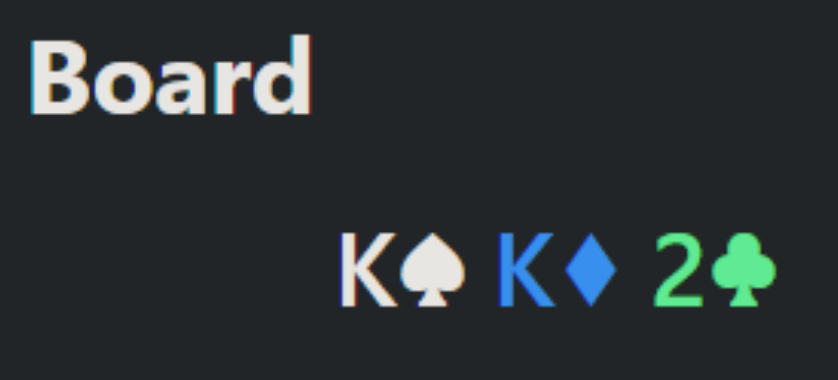}
        \caption{Paired Dry}
    \end{subfigure}
    \hfill
    \begin{subfigure}[b]{0.2\textwidth}
        \centering
        \includegraphics[width=\textwidth]{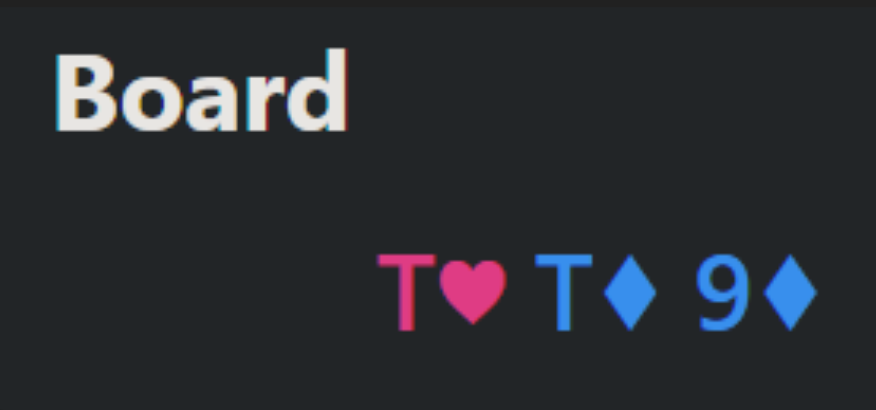}
        \caption{Paired Dynamic}
    \end{subfigure}
    \hfill

    \caption{Different board textures used to classify flops for filtering post-flop search space.}
    \label{fig:textrures}
\end{figure*}

The different board textures are shown in Figure \ref{fig:textrures}. With these textures, we present a comprehensive group of scenarios on the flop. Hands that fall in the above categories are usually played in very similar ways. The term ``\textit{wet}" is used to describe flops that have a likelihood of making the hands of some player. A ``\textit{dynamic}" board is one where the best hand may not be made yet but is likely to show up as the next two cards are dealt. A ``\textit{mid}" board represents medium-rank cards, whereas a ``\textit{low}" board represents a board with low-rank cards. A ``\textit{monotone}" board represents a board where all cards have the same suit.

\subsection{GPT-4 vs. Fine-tuned Model Gameplay Analysis}\label{sec: gameplay_analysis}

Pre-flop: As shown in Table \ref{tab:tournament-results-gpt4}, our fine-tuned Llama model performs better pre-flop. Through aggregating actions taken by both player from the game log, we find that Llama played remarkably more aggressive than GPT-4. Specifically, out of the 752 hands that end pre-flop
Llama open-raises (raises as the first player that moves) 27.3\% of the times and GPT-4 open-raises 15.3\% of the times. Compared to GTO pre-flop strategy, GPT-4 is playing too tightly, only raising the premium hands and thus losing overall by giving up their blinds most of the time. We see a similar pattern for situations with multiple raises pre-flop, where Llama is balancing between raising strong hands like KK and AKs with bluffing hands like KTo, while GPT-4 only reraises with AA,KK,AKs. GPT-4's lack of bluff in their raises presents an exploitable leak: if opponents know GPT-4 is never raising with weak hands, they will easily fold to GPT-4's raise, leading to GPT-4 missing out value when holding strong hands. 

Flop/Turn: Most of the winning of GPT-4 comes from hands that end flop or turn. After inspecting the specific game history, we find that one primary source of GPT-4 winning is adopting a ``donking" strategy, in which the out-of-position player who is not the pre-flop aggressor\footnote{Pre-flop aggressor refers to the player that raises the last before the flop. In general, the pre-flop aggressor would hold a stronger hand.} decides to bet as the flop/turn cards come out. In theory, donking gives negative expected value because pre-flop aggressor holds a stronger range of hands, including premium hands like AA,KK,QQ, while the non-aggressor holds a weaker range of hands. Betting a weaker range against a stronger range automatically leads to losing bigger pots with the same winning probabilities. Also, as mentioned previously, a key concept in GTO play is balance. It is generally very hard to balance donking with very strong hand with bluffs as now player needs to separate their range of hands into a more complicated game tree with donking available. Under specific board textures, this move can sometimes be adopted by GTO strategy, but we find GPT-4 was donking on spots that are clearly losing plays (not adopted by GTO strategy), if countered properly. However, as donking usually constitute a very small portion of a GTO strategy, and our dataset assumes GTO play by both players, it is not well-represented in our dataset and thus our fine-tuned model plays poorly against this strategy. Out of the 60 hands that GPT4 acts as the non-aggressor pre-flop, it chose to donk 15 hands and strikingly won \textbf{all} of them. 

Another main factor of GPT-4's winning is its aggressive postflop playing style. Concretely, GPT-4 on average bet/raised 0.823 times per street\footnote{A street refers to a game stage.} while Llama bet/raised 0.312 times. It is worth noticing that this over-aggressiveness is different from Llama's pre-flop aggression: Llama's pre-flop plays are closer to GTO strategies, while GPT-4's aggression is likely deviating from it. Nevertheless, Llama is also not playing optimally, instead deviating to the passive side, folding more than optimal plays. This coincidentally results in a style suppression. A good example of this can be illustrated by rock-paper-scissors games. Imagine player A chooses rock 100\% of time, while player B chooses scissors 50\% of time and rock 50\% of time. We can see that even though player B has a strategy closer to the optimal, it will never win against player A's more suboptimal strategy. Similarly, while GPT-4 has a strategy farther from GTO (as assessed by our dataset), it is still able to win against Llama. 

River: The action distribution for hands in the river aligns with the style analysis in flop/turn. Out of the 80 hands, GPT-4 bet/raised 0.465 times per street in average, while Llama bet/raised 0.134 times only. However, this time, because of the over-aggression of GPT-4 on flop and turn, only Llama's very strongest hands remains in the river and so it easily wins the river by catching bluffs from GPT-4 using premium hands made postflop.

\begin{table}
\scalebox{0.90}{
    \centering
    \begin{tabular}{p{1.8cm}|p{2.5cm}|p{1.5cm}|p{1.5cm}} 

    \centering \textbf{GAME STAGE} & \textbf{ACTION} & \textbf{GPT-4} & \textbf{LLAMA}\\ \hline\hline
    \centering Pre-flop & Open-Raises (Percentage)  & 15.3\% & \textbf{27.3\%} \\ \hline
    \centering Flop + Turn & Donking-Bets (Percentage)  & \textbf{25.0\%} & 0.0\% \\ \hline
    \centering Flop + Turn & Bets/Raises (Times/Street) & \textbf{0.823} & 0.312 \\ \hline
    \centering River & Bets/Raises (Times/Street) & \textbf{0.465} & 0.134 \\ \hline
    \end{tabular}
}

    \caption{Playing Style Summary of GPT-4 vs. Player 5000}
    \label{tab:gameplay_freq_analysis}
\end{table}

\begin{table*}[ht]
\centering
\begin{tabular}{{p{15cm}p{0.1cm}}}

\textbf{Few-Shot Prompt}&\\ \hline

\texttt{You are a specialist in playing 6-handed No Limit Texas Holdem. The following will be a game scenario and you need to make the optimal decision. 
\newline
\newline
Here is a game summary: 
\newline
\newline
The small blind is 0.5 chips and the big blind is 1 chips. Everyone started with 100 chips. 
\newline
The player positions involved in this game are UTG, HJ, CO, BTN, SB, BB.
\newline
In this hand, your position is HJ, and your holding is [King of Heart and King of Spade].
\newline
Before the flop, HJ raise 2.0, CO raise 6.5, and SB raise 17.5. Assume that all other players that is not mentioned folded.
\newline
\newline
Now it is your turn to make a move. 
\newline
To remind you, the current pot size is 27.0 chips, and your holding is [King of Heart and King of Spade].
\newline
\newline
Decide on an action based on the strength of your hand on this board, your position, and actions before you. Do not explain your answer.
\newline
Your optimal action is: \textbf{all in}}

\end{tabular}
\caption{Example prompt used for few-shot evaluations of pre-trained LLMs. The above example is a prompt for one single spot. In the few-shot scenario, multiple such spots are presented to the model before the query spot is presented. }
\label{table:prompt}
\end{table*}

\begin{figure}
    \centering
    
    \centering
    \includegraphics[width=0.5\textwidth]{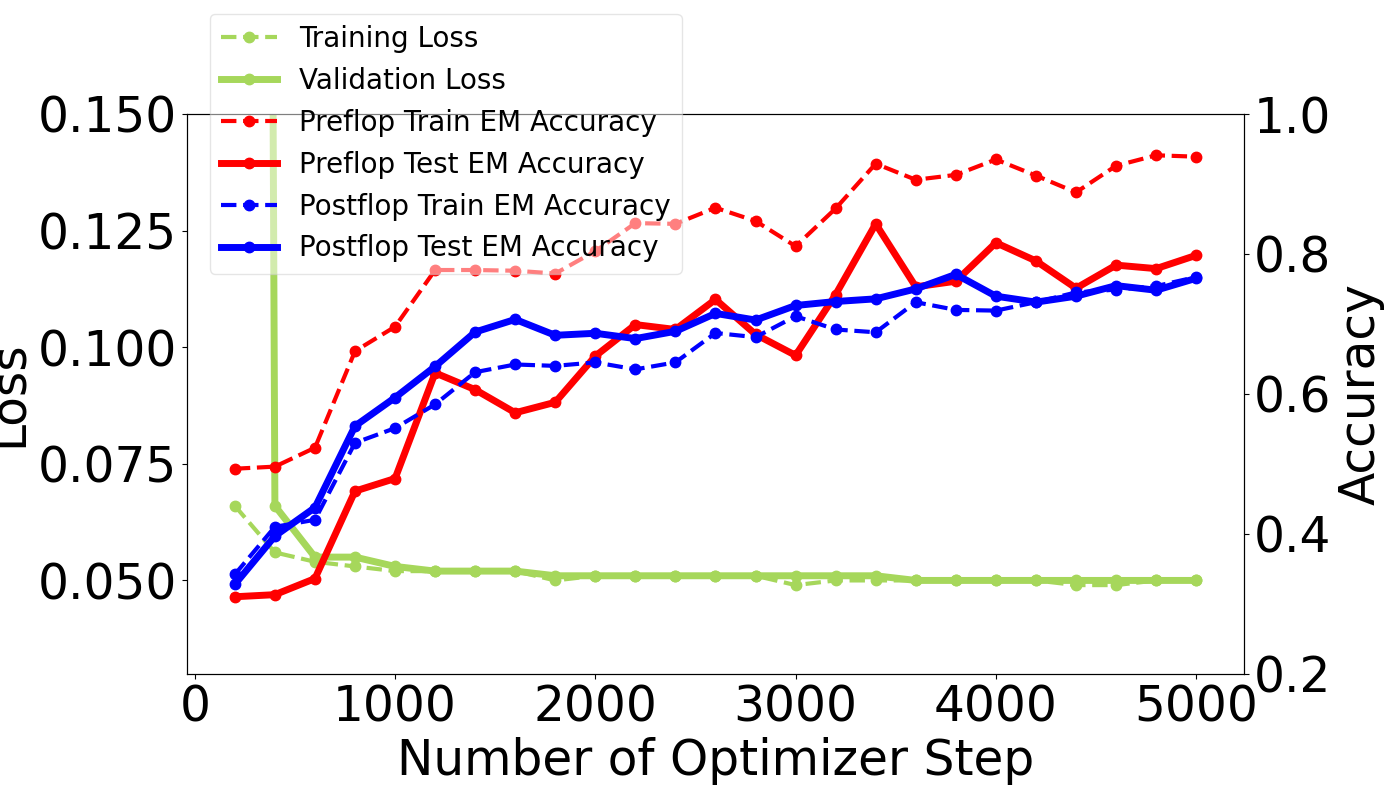}
    \caption{Training dynamics of Llama-2-7B on \textsc{PokerBench} training dataset.}\label{fig:sft-llama-2}
\end{figure}

\begin{figure}
    \centering
    
    \centering
    \includegraphics[width=0.5\textwidth]{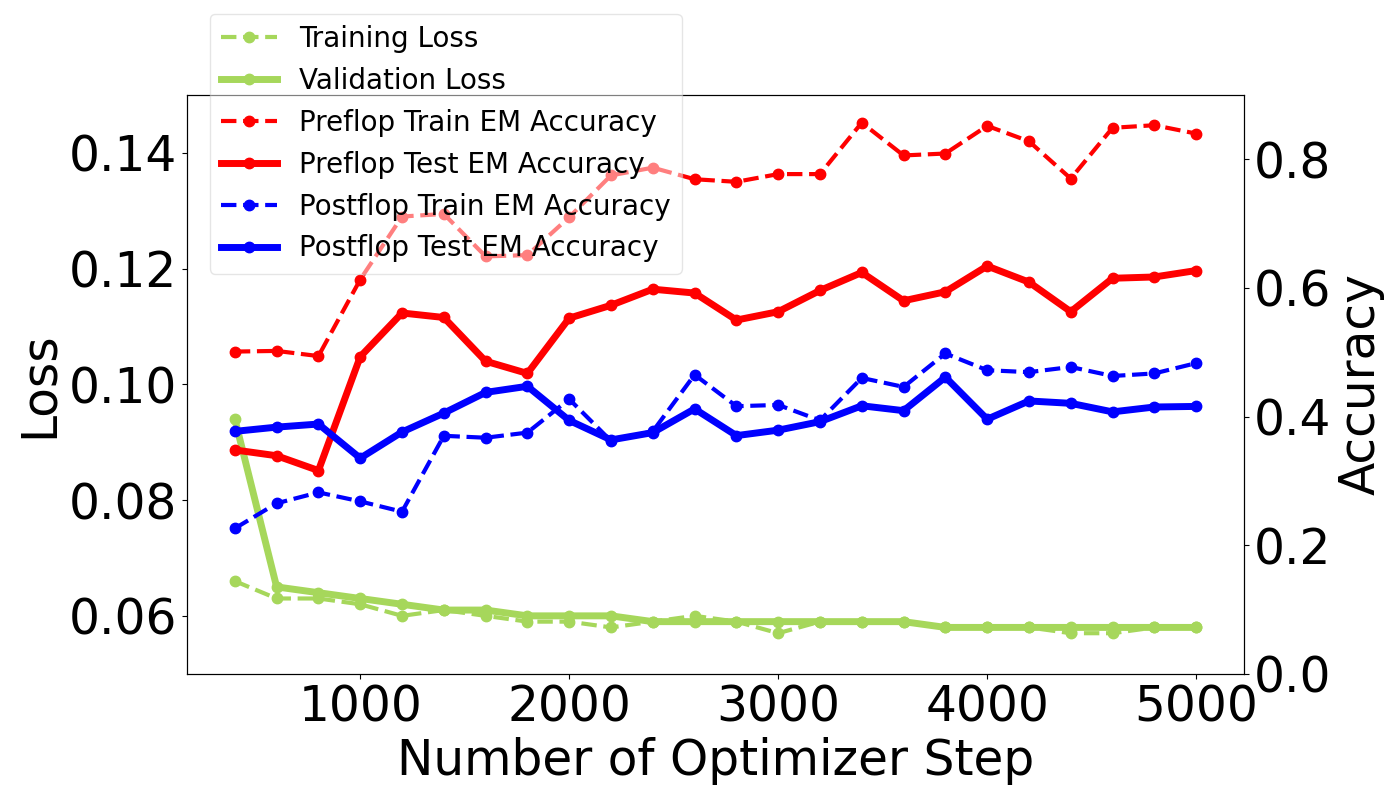}
    \caption{Training dynamics of Gemma-2B on \textsc{PokerBench} training dataset.}\label{fig:sft-gemma}
\end{figure}

\end{document}